\documentclass{article}
\usepackage{ijcai25}

\usepackage{times}
\usepackage{soul}
\usepackage{url}
\usepackage[hidelinks]{hyperref}
\usepackage[utf8]{inputenc}
\usepackage[small]{caption}
\usepackage{graphicx}
\usepackage{amsmath}
\usepackage{amsthm}
\usepackage{booktabs}
\usepackage{algorithm}
\usepackage{algorithmic}
\usepackage[switch]{lineno}
\usepackage{multirow}
\usepackage{amssymb}
\usepackage{placeins}
\hypersetup{hypertex=true,
            colorlinks=true,
            linkcolor=red,
            anchorcolor=red,
            citecolor=black}

\title{METOR: A Unified Framework for Mutual Enhancement of Objects and Relationships in Open-vocabulary Video Visual Relationship Detection}

\author{
    Yongqi~Wang$^{1,2}$\and
    Xinxiao~Wu$^{1,2}$\and
    Shuo~Yang$^2$\thanks{Corresponding author.}\\
    \affiliations
    \textsuperscript{\rm 1}Beijing Key Laboratory of Intelligent Information Technology, School of Computer Science \& Technology, Beijing Institute of Technology, China\\
    \textsuperscript{\rm 2}Guangdong Laboratory of Machine Perception and Intelligent Computing, Shenzhen MSU-BIT University, China\\
    \emails
    \{3120230916, wuxinxiao\}@bit.edu.cn,
    yangshuo@smbu.edu.cn
}

\begin{document}

\maketitle

\begin{abstract}
    Open-vocabulary video visual relationship detection aims to detect objects and their relationships in videos without being restricted by predefined object or relationship categories.
    Existing methods leverage the rich semantic knowledge of  pre-trained vision-language models such as CLIP to identify novel categories. They typically adopt a cascaded pipeline to first detect objects and then classify relationships based on the detected objects, which may lead to error propagation and thus suboptimal performance.
    In this paper, we propose \textbf{M}utual \textbf{E}nhancemen\textbf{T} of \textbf{O}bjects and \textbf{R}elationships (METOR), a query-based unified framework to jointly model and mutually enhance object detection and relationship classification in open-vocabulary scenarios.
    Under this framework, we first  design a CLIP-based contextual refinement encoding module that extracts visual contexts of objects and relationships  to refine the encoding of text features and object queries, thus improving the  generalization of encoding to novel categories. Then we propose an iterative enhancement module to alternatively enhance the representations of objects and relationships  by fully exploiting their interdependence to improve recognition performance.
    Extensive experiments on two public datasets, VidVRD and VidOR, demonstrate that our framework achieves state-of-the-art performance.
    Codes are at \url{https://github.com/wangyongqi558/METOR}.
\end{abstract}

\begin{figure}[t]
    \centering
    \includegraphics[width=1\linewidth]{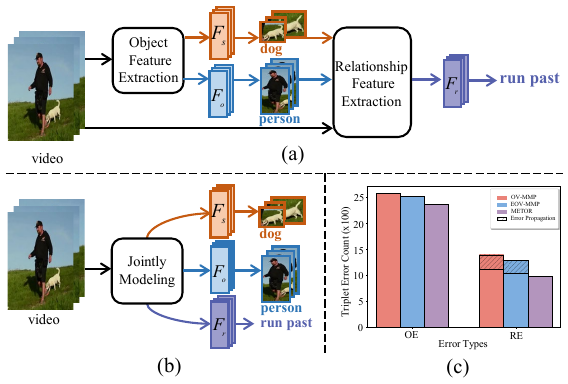}
    \caption{(a) Existing methods adopt a cascaded pipeline. (b) Our method jointly models objects and relationships for their mutually enhancement. (c) Statistics of triplet errors on the VidVRD test set, categorized into object errors (OE) and relationship errors (RE). 
    We also highlight the relationship errors caused by error propagation in the cascaded methods (OV-MMP and EOV-MMP), where the relationship is correctly classified using ground-truth object trajectories but misclassified using detected object trajectories.}
    \label{fig:intro}
    \end{figure}

\section{Introduction}

Open-vocabulary video visual relationship detection (Open-VidVRD)~\cite{gao2023compositional} focuses on detecting relationships between objects in videos, typically represented as triplets of the form $\langle subject, relationship, object \rangle$, following an open-vocabulary setting.
In this setting, both object and relationship categories are divided into a base set and a novel set.
The model is trained on the base set and is expected to generalize to  the novel set during testing, enabling relationship detection in a wider range of open-vocabulary scenarios.

Recent advances in  pre-trained vision-language models~\cite{radford2021learning,li2022align,li2023blip} have demonstrated significant potential in enhancing open-vocabulary tasks~\cite{liu2024open,wu2024building,fan2024active}. By leveraging vast amounts of vision-language pairs during training, these models effectively encode rich semantic knowledge encompassing entities, actions, scenes, and relationships~\cite{fang2024simple,wang2021actionclip,liang2023open,gao2023compositional}. 
In the Open-VidVRD task, these pre-trained models have been used to recognize novel object and relationship categories, thereby improving the generalization capability beyond predefined categories.
Existing Open-VidVRD methods~\cite{gao2023compositional,yang2024multi,wu2024open,wang2024end} typically adopt a cascaded pipeline, first detecting objects and then classifying relationships based on the detected objects, as illustrated in Fig.~\ref{fig:intro} (a).
This design often leads to error propagation, as the inaccuracy of object detection adversely affects the relationship classification, resulting in suboptimal performance.
Fig.\ref{fig:intro} (c) shows the number and causes of triplet errors for different methods on the VidVRD dataset. It is evident that the cascaded methods exhibit a significant increase in relationship errors due to error propagation, underscoring the inherent weaknesses of these methods.

In this paper, we propose \textbf{M}utual \textbf{E}nhancemen\textbf{T} of \textbf{O}bjects and \textbf{R}elationships (METOR), a query-based unified framework for Open-VidVRD, which jointly models and mutually enhances object detection and relationship classification, as shown in Fig.~\ref{fig:intro} (b). It simplifies the Open-VidVRD process by adopting a unified modeling strategy, thereby mitigating the error propagation inherent in cascaded pipelines, and emphasizes the role of relationship context in enriching object representations, effectively exploiting the interdependence between objects and relationships to promote their mutual enhancement. As illustrated in Fig.~\ref{fig:intro} (c), our method reduces both object errors and relationship errors, highlighting the benefits of the proposed framework.

To enhance the generalization ability to novel categories, we design a CLIP-based contextual refinement encoding module that captures the contextual information of objects and relationships to refine the encoding process. 
Specifically, we incorporate learnable object and relationship tokens into the CLIP visual encoder to capture their respective contexts. 
These contexts are then used to refine the CLIP-encoded text features and object queries, delivering instance-specific semantic knowledge to improve the adaptability in open-vocabulary scenarios.

To fully exploit the interdependence between objects and relationships for recognition, we propose an iterative enhancement module to alternately enhance the representations of objects and relationships. Specifically, this module consists of multiple iterative enhancement layers, where each layer first uses object features to extract relationship features through spatio-temporal modeling and then uses the extracted relationship features to refine the object features, promoting a continuous mutual enhancement process that enables objects and relationships to iteratively improve each other's representations.

To summarize, the main contributions are as follows:
\begin{itemize}
    \item We propose METOR, a query-based unified framework that jointly models object detection and relationship classification to effectively exploit their interdependence to promote mutual enhancement, simplifying the process of Open-VidVRD.
    \item We propose an iterative enhancement module that alternately enhances the representations of objects and relationships by using each other's representations for more accurate recognition.
    \item We design a CLIP-based contextual refinement encoding module that extracts contexts for objects and relationships to refine the encoding of text features and object queries for better open-vocabulary generalization.

\end{itemize}

\begin{figure*}[t]
\centering
\includegraphics[width=1\textwidth]{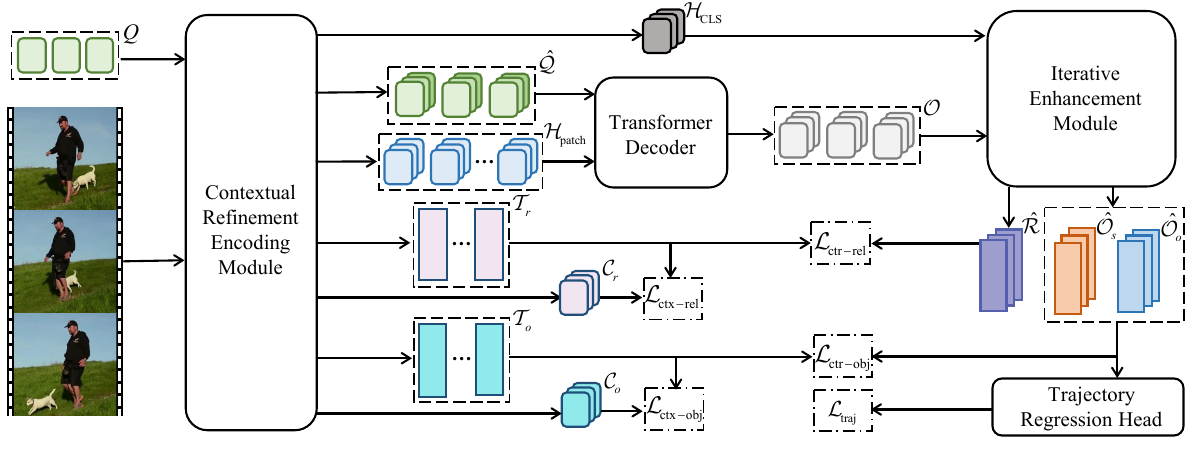}
\caption{
Overview of the proposed framework.}
\label{fig:overview}
\end{figure*}

\section{Related Work}

\textbf{Video visual relationship detection (VidVRD)}~\cite{shang2017video} aims to detect relationships between objects over time in a given video, which has been widely applied to various visual understanding tasks~\cite{zhao2023constructing,nguyen2024hig,rodin2024action}.
Numerous studies have explored various VidVRD methods, which can be broadly categorized into spatio-temporal modeling~\cite{qian2019video,tsai2019video,liu2020beyond,cong2021spatial}, relationship refinement~\cite{shang2021video,chen2021social}, video relationship debiasing~\cite{xu2022meta,dong2022stacked,lin2024td2}, and end-to-end video relationship detection~\cite{zheng2022vrdformer,zhang2023end,jiang2024vrdone}.
However, these methods are designed for close-set scenarios and struggle to generalize to open-vocabulary settings, thus being limited in pratical applications.

\textbf{Open-vocabulary VidVRD (Open-VidVRD)}~\cite{gao2023compositional} has emerged in recent years to  extend VidVRD by testing novel categories, thereby enhancing its applicability to real-world scenarios.
RePro~\cite{gao2023compositional}, OV-MMP~\cite{yang2024multi} and UASAN~\cite{wu2024open} design prompt learning or semantic alignment modules to better align visual and textual modalities.
However, these methods rely on a close-set pre-trained trajectory detector for trajectory detection.
Incorrect trajectories can lead to relationship classification errors, causing severe error propagation and hurting the overall performance.

The method most related to ours is EOV-MMP~\cite{wang2024end}, which extends OV-MMP into an end-to-end model, eliminating the need for the close-set pre-trained trajectory detector used in previous Open-VidVRD methods. 
It jointly optimizes the object detection and relationship classification modules, introducing an auxiliary loss to capture relationship context during object detection.
However, EOV-MMP only perceives the relationship context without explicitly leveraging it to enhance object representations. In addition, it still follows a cascaded pipeline that first detects objects and then classifies relationships. 
In contrast, our METOR jointly models objects and relationships, and fully exploits their interdependence to promote mutual enhancement.

\section{Our Framework}
\subsection{Problem Definition}
Video Visual Relationship Detection (VidVRD) involves identifying visual relationship instances in a video sequence \( \mathcal{V} = \{f_t\}_{t=1}^{T} \), where \( f_t \) represents the frame at timestamp \( t \), and \( T \) denotes the total number of frames. Each visual relationship instance is represented as a tuple \( (s, r, o, \tau_s, \tau_o) \), where \( s \), \( r \), and \( o \) indicate the subject, relationship, and object categories, respectively.
\( \tau_s \) and \( \tau_o \) represent the trajectories of  subject and object, respectively, defined as sequences of bounding boxes \( \{b_t^s\} \) and \( \{b_t^o\} \) over a temporal span, respectively, where \( t \) ranges from \( t_{\text{start}} \) to \( t_{\text{end}} \), indicating the start and end times of each trajectory.
In Open-VidVRD, categories are divided into base and novel splits, including base object categories $\mathbb{C}_o^b$, novel object categories $\mathbb{C}_o^n$,  base relationship categories $\mathbb{C}_r^b$,  and novel relationship categories $\mathbb{C}_r^n$. Training is only performed  on base categories, while evaluation includes both base and novel categories to evaluate the model's generalization.

\subsection{Overview}
In this paper, we propose METOR, a query-based unified framework that jointly models object detection and relationship classification to effectively exploit their interdependence to promote mutual enhancement, streamlining the process of Open-VidVRD.
The framework comprises two key modules: 
a contextual refinement encoding module (Sec.~\ref{sec:enc}) and an iterative enhancement module (Sec.~\ref{sec:iter}).
An overview of METOR is illustrated in Fig.~\ref{fig:overview}.

For a given video \( \mathcal{V} \), we input it into the contextual refinement encoding module along with $N_q$ learnable object queries $\mathcal{Q}$.
This module outputs CLS embeddings $\mathcal{H}_{\text{CLS}}$, patch embeddings $\mathcal{H}_{\text{patch}} $, refined text features \( \mathcal{T}_o \) for objects, refined text features  \( \mathcal{T}_r \) for relationships, refined object queries \( \hat{\mathcal{Q}} \), object context embeddings  $\mathcal{C}_o $ and relationship context embeddings $\mathcal{C}_r $, formulated by
\begin{equation}
(\mathcal{H}_{\text{CLS}},\mathcal{H}_{\text{patch}},\mathcal{T}_o,\mathcal{T}_r,\hat{\mathcal{Q}},\mathcal{C}_o,\mathcal{C}_r)=\Phi(\mathcal{V},\mathcal{Q}),
\end{equation}
where $\Phi(\cdot)$ denotes the contextual refinement encoding module. 

Then, the refined object queries and patch embeddings are then passed through a Transformer decoder to generate visual object features:
\begin{equation}
\mathcal{O}=Decoder(\hat{\mathcal{Q}},\mathcal{H}_{\text{patch}}),
\end{equation}
where $Decoder(\cdot)$ is the Transformer decoder, and $\mathcal{O}$ denotes the visual features of $N_q$ objects in the video.

Next, the visual object features and the CLS embeddings containing global semantic information are fed into the iterative enhancement module to generate mutually enhanced visual subject feature $\hat{\mathcal{O}}_s $, visual object feature $\hat{\mathcal{O}}_o $, and visual relationship feature $\hat{\mathcal{R}} $:
\begin{equation}
(\hat{\mathcal{O}}_s,\hat{\mathcal{O}}_o,\hat{\mathcal{R}})=\Psi(\mathcal{O},\mathcal{H}_{\text{CLS}}),
\end{equation}
where $\Psi(\cdot)$ denotes the iterative enhancement module.

Finally, the mutually enhanced features are  matched with the corresponding textual features  to predict subject score $\mathcal{S}_s$, object score $\mathcal{S}_o$, and relationship score $\mathcal{S}_r$:
\begin{equation}
\begin{gathered}
\mathcal{S}_s = \sigma(\cos(\hat{\mathcal{O}}_s,\mathcal{T}_o)), \\
\mathcal{S}_o = \sigma(\cos(\hat{\mathcal{O}}_o,\mathcal{T}_o)), \\
\mathcal{S}_r = \sigma(\cos(\hat{\mathcal{R}},\mathcal{T}_r)),
\label{eq:cos}
\end{gathered}
\end{equation}
where $\sigma(\cdot)$ is the sigmoid function, and $\cos(\cdot, \cdot)$ represents the cosine similarity.
The trajectories for the subject and object are predicted as
\begin{equation}
\begin{gathered}
\tau_s = {M_b}(\hat{\mathcal{O}}_s), \\
\tau_o = {M_b}(\hat{\mathcal{O}}_o),
\label{eq:box}
\end{gathered}
\end{equation}
where ${M_b}(\cdot)$ denotes a trajectory regression head implemented as a multi-layer perceptron.

\begin{figure*}[t]
\centering
\includegraphics[width=1\textwidth]{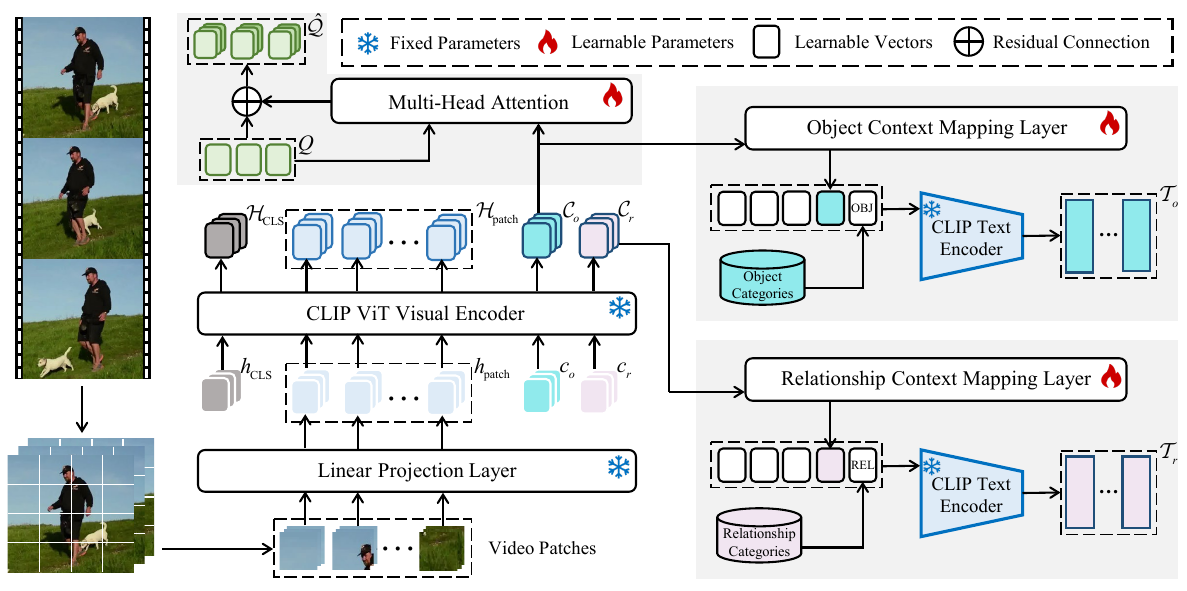}
\caption{
An overview of the proposed contextual refinement encoding module.}
\label{fig:encoding}
\end{figure*}

\subsection{Contextual Refinement Encoding Module}
\label{sec:enc}
The main goal of Open-VidVRD is to discover novel categories. 
To improve the generalization ability to novel categories, we propose a contextual refinement encoding module that extracts contexts for objects and relationships to refine the encoding of text features and object queries.
An illustration of this module is shown in Fig.~\ref{fig:encoding}.

The video \( \mathcal{V} \) is first divided into fixed-sized and non-overlapping patches, which are then linearly projected into 1D tokens:
\begin{equation}
{h}_{\text{patch}} = L(P(\mathcal{V})),
\end{equation}
where \( P(\cdot) \) represents the patchification operation, and $L(\cdot)$ is the linear projection layer.
\( h_{\text{patch}} \in \mathbb{R}^{T \times N_p \times d} \) represents the  patch tokens, where \( T \) is the number of video frames, and \( N_p \) is the number of patches per frame.

The patch tokens, concatenated with learnable CLS tokens \( h_{\text{CLS}}\), learnable object context tokens $c_o$ and learnable relationship context tokens $c_r$, are fed into the CLIP ViT visual encoder to generate CLS embeddings $\mathcal{H}_{\text{CLS}}$, patch embeddings $\mathcal{H}_{\text{patch}}$, object context embeddings $\mathcal{C}_o$ and relationship context embeddings $\mathcal{C}_r$:
\begin{equation}
\begin{gathered}
    ({\mathcal{H}}_{\text{CLS}},{\mathcal{H}}_{\text{patch}},{\mathcal{C}}_o,{\mathcal{C}}_r)=V([h_{\text{CLS}}; h_{\text{patch}}; c_o; c_r]),
\label{equ:alpha}
\end{gathered}
\end{equation}
where \( V(\cdot) \) represents the CLIP ViT visual encoder, and $[\cdot;\cdot]$ denotes the concatenation operation.

To capture contextual dependencies to refine the object queries in each frame, the object context embedding are used as keys and values, and the object queries serve as queries in a multi-head attention mechanism:  
\begin{equation}
\hat{\mathcal{Q}} = \text{MHA}(\mathcal{Q}, \mathcal{C}_o, \mathcal{C}_o) + \mathcal{Q},
\end{equation}
where \( \text{MHA}(\cdot,\cdot,\cdot) \) is the multi-head attention mechanism. The residual connection preserves information in the original object queries while incorporating contextual knowledge from the object context embeddings.

To refine the textual object features, the object context embeddings are first processed through an object context mapping layer, then concatenated with a set of learnable vectors \(\mathbf{v}_o\) and an object category embedding \text{OBJ}, and finally passed through the CLIP text encoder:
\begin{equation}
\mathcal{T}_o = G([ \mathbf{v}_o;\mathcal{M}_{o}({\mathcal{C}}_o);\text{OBJ}]),
\end{equation}
where $\mathcal{T}_o$ denotes the refined textual object features,  $G(\cdot)$ represents the CLIP text encoder, and $\mathcal{M}_o$ denotes the object context mapping layer, implemented as a multi-layer perceptron. 

In a similar way, the refined textual relationship features \(\mathcal{T}_r\) are learned from the relationship context embeddings \(\mathcal{C}_r\):
\begin{equation}
\begin{gathered}
   \mathcal{T}_r = G([ \mathbf{v}_r;\mathcal{M}_{r}({\mathcal{C}}_r);\text{REL}]),
\label{equ:beta}
\end{gathered}
\end{equation}
where $\mathbf{v}_r$ denotes a set of learnable vectors, $\mathcal{M}_r$ represents the relationship context mapping layer, implemented as a multi-layer perceptron, and $\text{REL}$ corresponds to the relationship category embedding.

After encoding, the refined object queries $\hat{\mathcal{Q}}$ and patch embeddings $\mathcal{H}_{\text{patch}}$ are fed into a Transformer decoder to generate visual object features $\mathcal{O} \in \mathbb{R}^{N_q \times T \times d}$, which represent the visual features corresponding to $N_q$ candidate objects across $T$ video frames. From the visual object features \(\mathcal{O}\), we derive subject-object pairs for subsequent  recognition, denoted as \((\mathcal{O}_s, \mathcal{O}_o)\), where \(\mathcal{O}_s\) and \(\mathcal{O}_o\) represent the visual subject and object features, respectively. 

\subsection{Iterative Enhancement Module}
\label{sec:iter}

\begin{figure}
    \centering
    \includegraphics[width=1\linewidth]{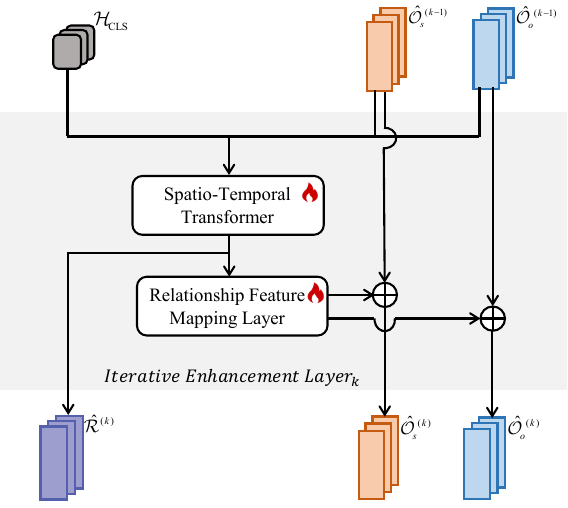}
    \caption{An overview of the proposed iterative enhancement module.}
    \label{fig:iter}
    \end{figure}

We propose an iterative enhancement module to alternatively enhance the representations of objects and relationships by fully exploiting their interdependence, which consists of  \( N_i \) iterative enhancement layers,  as illustrated in Fig.~\ref{fig:iter}.
In each layer, the visual features of the paired subject and object are first concatenated with the CLS embeddings,  and then is fed into a spatio-temporal Transformer  to generate visual relationship features. These visual relationship features are processed through a relationship feature mapping layer, which in turn enhances the visual subject and object features. This process is expressed as
\begin{equation}
\begin{gathered}
   \hat{\mathcal{R}}^{(k)} = STTrans^{(k)}([\hat{\mathcal{O}}_s^{(k-1)}; \hat{\mathcal{O}}_o^{(k-1)}; \mathcal{H}_{\text{CLS}}]),\\
   \hat{\mathcal{O}}_s^{(k)} =\alpha \hat{\mathcal{O}}_s^{(k-1)} + (1-\alpha)M_f^{(k)}(\hat{\mathcal{R}}^{(k)}),\\
   \hat{\mathcal{O}}_o^{(k)} =\alpha \hat{\mathcal{O}}_o^{(k-1)} + (1-\alpha)M_f^{(k)}(\hat{\mathcal{R}}^{(k)}),
\label{equ:beta}
\end{gathered}
\end{equation}
where \( STTrans^{(k)}(\cdot) \) and \( M_f^{(k)}(\cdot) \) denote the spatio-temporal Transformer and the relationship feature mapping layer of the \( k \)-th iterative enhancement layer, respectively. $\hat{\mathcal{R}}^{(k)}$ represents the visual relationship features, while \(\hat{\mathcal{O}}_s^{(k)}\) and \(\hat{\mathcal{O}}_o^{(k)}\) are the enhanced visual subject and object features in the \(k\)-th layer.
 \(\alpha \) is a balance parameter. 

For a subject-object feature pair $(\mathcal{O}_s,\mathcal{O}_o)$, the iterative enhancement module initializes the inputs to the first layer as $\hat{\mathcal{O}}_s^{(0)}=\mathcal{O}_s$ and $\hat{\mathcal{O}}_o^{(0)}=\mathcal{O}_o$, and outputs the the mutually enhanced visual subject, object and relationship features as
 \(\hat{\mathcal{O}}_s=\hat{\mathcal{O}}_s^{(N_i)}\), \(\hat{\mathcal{O}}_o=\hat{\mathcal{O}}_o^{(N_i)}\), and \(\hat{\mathcal{R}}=\hat{\mathcal{R}}^{(N_i)}\), respectively.

\subsection{Training Objective}
We train the entire framework in an end-to-end manner.  
The overall objective function consists of five losses: a relationship contrastive loss \(\mathcal{L}_{\text{rel-ctr}}\), an object contrastive loss \(\mathcal{L}_{\text{obj-ctr}}\), a trajectory loss \(\mathcal{L}_{\text{traj}}\), a relationship contextual loss \(\mathcal{L}_{\text{rel-ctx}}\), and an object contextual loss \(\mathcal{L}_{\text{obj-ctx}}\), formulated by 
\begin{equation}
    \mathcal{L} = \mathcal{L}_{\text{rel-ctr}} + \mathcal{L}_{\text{obj-ctr}}\ + \theta_{\text{traj}} \mathcal{L}_{\text{traj}} + \theta_{\text{ctx}} (\mathcal{L}_{\text{rel-ctx}} + \mathcal{L}_{\text{obj-ctx}}),
\label{equ:twop}
\end{equation}
where \(\theta_{\text{traj}}\) and \(\theta_{\text{ctx}}\) are balance factors.

\noindent \textbf{Contrastive Loss.}
The relationship contrastive loss is formulated using the binary cross-entropy loss (BCE):
\begin{equation}
    \mathcal{L}_{\text{rel-ctr}} = \frac{1}{|\mathbb{C}_r^b|}{\rm BCE}(\mathcal{S}_r, \tilde{r}),
\end{equation}
where \(\mathcal{S}_r\) represents the predicted relationship score, \(\tilde{r}\) is the ground-truth relationship labels for the subject-object pair, and \(\mathbb{C}_r^b\) denotes the set of base relationship categories used during training. 

The object contrastive loss is computed using the cross-entropy loss (CE):
\begin{equation}
    \mathcal{L}_{\text{obj-ctr}} = {\rm CE}(\mathcal{S}_s, \tilde{s})+{\rm CE}(\mathcal{S}_o, \tilde{o}), 
\end{equation}
where \(\mathcal{S}_s\) and \(\mathcal{S}_o\) are the predicted scores for the subject and object, respectively, and \(\tilde{s}\) and \(\tilde{o}\) are the ground-truth labels for the subject and object categories. 

\noindent \textbf{Trajectory Loss.}
The trajectory loss consists of a bounding box regression loss and a trajectory consistency loss, given by 
\begin{equation}
\begin{gathered}
    \mathcal{L}_{\text{traj}} = \mathcal{L}_{\text{box}} + \theta_{\text{cst}} \mathcal{L}_{\text{cst}}, \\
    \mathcal{L}_{\text{box}} = \frac{1}{|T|} \sum\nolimits_{t=1}^{T} \sum\nolimits_{e \in \{s, o\}} \text{SL1}(\mathcal{B}_e^{(t)}, \tilde{\mathcal{B}}_e^{(t)}), \\
    \mathcal{L}_{\text{cst}} = \frac{1}{|T-1|} \sum\nolimits_{t=1}^{T-1} \sum\nolimits_{e \in \{s, o\}} \|\mathcal{B}_e^{(t+1)} - \mathcal{B}_e^{(t)}\|_1,
\label{equ:onep}
\end{gathered}
\end{equation}
where \(\mathcal{B}_e^{(t)}\) and \(\tilde{\mathcal{B}}_e^{(t)}\) denote the predicted and ground truth bounding boxes for entity \( e \in \{s, o\} \) at frame \( t \), derived from the predicted trajectory \(\tau_e\) and ground truth trajectory \(\tilde{\tau}_e\).
\(\text{SL1}(\cdot)\) represents the Smooth L1 loss for bounding box regression. \(\|\cdot\|_1\) is the L1 norm enforcing temporal smoothness by penalizing large deviations between consecutive bounding boxes.
\(\theta_{\text{cst}}\) balances the contributions of the bounding box regression loss and the trajectory consistency loss.

\noindent \textbf{Contextual Loss.}
To capture more effective contextual information for relationships and objects, we define the  contextual losses for them using  the binary cross-entropy loss (BCE):
\begin{equation}
\begin{gathered}
\mathcal{L}_{\text{rel-ctx}} = \text{BCE}(\cos(\mathcal{C}_r, \mathcal{T}_r), \tilde{\mathcal{R}}),\\
\mathcal{L}_{\text{obj-ctx}} = \text{BCE}(\cos(\mathcal{C}_o, \mathcal{T}_o), \tilde{\mathcal{O}}),
\end{gathered}
\end{equation}
where \(\tilde{\mathcal{R}}\) and \(\tilde{\mathcal{O}}\) represent the sets of relationship and object categories present in each frame, respectively.

\begin{table*}
\centering
\begin{tabular}{c|c|cccc|cccc} 
\hline \hline
\multirow{2}{*}{Split} & \multirow{2}{*}{Method} & \multicolumn{4}{c|}{VidVRD} & \multicolumn{4}{c}{VidOR}\\
\cline{3-10}
\multicolumn{1}{c|}{}&\multicolumn{1}{c|}{}&mAP&R@50&R@100&mAP$_o$&mAP&R@50&R@100&mAP$_o$\\
\hline
\multirow{5}{*}{Novel}
& RePro~\cite{gao2023compositional}
&5.87&12.75&16.23&10.36&-&-&-&-\\
&UASAN~\cite{liu2024open}
&9.88&12.80&17.68&12.15&-&-&-&-\\
&OV-MMP~\cite{yang2024multi}
&12.15&13.72&15.21&14.37&0.84&1.44&1.44&1.11\\
&EOV-MMP~\cite{wang2024end}
&{15.04}&{16.03}&{18.18}&{36.31}&{2.45}&{4.79}&{4.79}&{2.33}\\
&METOR (Ours)&\textbf{16.74}&\textbf{16.72}&\textbf{19.43}&\textbf{38.91}&\textbf{3.75}&\textbf{4.86}&\textbf{5.32}&\textbf{3.37}\\
\hline
\hline
\multirow{5}*{All}
&RePro~\cite{gao2023compositional}
&21.12&12.63&15.42&18.18&-&-&-&-\\
&UASAN~\cite{liu2024open}
&22.93&15.74&18.89&23.74&-&-&-&-\\
&OV-MMP~\cite{yang2024multi}
&22.10&13.26&16.08&34.61&7.15&6.54&8.29&3.38\\
&EOV-MMP~\cite{wang2024end}
&{26.34}&{16.48}&{19.54}&{52.72}&{11.08}&{8.43}&\textbf{9.82}&{12.99}\\
&METOR (Ours)&\textbf{27.52}&\textbf{16.69}&\textbf{19.58}&\textbf{55.09}&\textbf{12.32}&\textbf{8.54}&{9.72}&\textbf{14.02}\\
\hline \hline
\end{tabular}
\caption{Comparison with existing methods on VidVRD and VidOR datasets.}
\label{tab:sota-vidvrd}
\end{table*}

\section{Experiment}
\subsection{Dataset and Evaluation}
\noindent \textbf{Datasets.}
We evaluate our framework on the {VidVRD}~\cite{shang2017video} and {VidOR}~\cite{shang2019relation} datasets. VidVRD consists of 1,000 videos, with 800 videos for training and 200 for testing, covering 35 object categories and 132 relationship categories. VidOR contains 10,000 videos, including 7,000 for training, 835 for validation, and 2,165 for testing, covering 80 object categories and 50 relationship categories.

\noindent \textbf{Evaluation Settings.}
Following Repro~\cite{gao2023compositional}, we designate common object and relationship categories as base categories and  rarer ones  as novel categories. Training is performed on the base categories, and testing is conducted under two settings: (1) {Novel-split} evaluation, which includes all object categories and novel relationship categories; (2) {All-split} evaluation, covering all object and relationship categories as the standard evaluation. Testing is carried out on both the VidVRD test set and the VidOR validation set, as the annotations for the VidOR test set are not available.

\noindent \textbf{Evaluation Tasks.}
Three evaluation tasks are usually used for VidVRD: scene graph detection ({SGDet}), scene graph classification ({SGCls}), and predicate classification ({PredCls}).
SGCls and PredCls are often used to evaluate  methods that depend on pre-detected trajectories and are not suitable for our framework which jointly models objects and relationships. So we use  SGDet that do not rely on pre-detected trajectories for evaluation.

\noindent \textbf{Evaluation Metrics.}
We use mean Average Precision ({mAP}) and Recall@K ({R@K}) with K = 50, 100 as evaluation metrics for relationship classification. Following EOV-MMP~\cite{wang2024end}, we also use mean Average Precision of object trajectory ({mAP$_o$}) to evaluate the quality of object detection.

\subsection{Implementation Details}

In all experiments, key frames are sampled every 30 video frames to form 30-frame video segments. Following~\cite{shang2017video,gao2023compositional,liu2024open}, visual relationship triplets are generated for video segments and merged using the greedy relation association algorithm proposed in~\cite{shang2017video}.
We adopt the ViT-L/14 variant of CLIP with fixed parameters.
The number of iterative enhancement layers $N_i$ is set to two for VidVRD and three for VidOR. The hyperparameter \(\alpha\) in Eq.~\ref{equ:beta} is set to 0.9. The loss balance factors \(\theta_{\text{traj}}\), \(\theta_{\text{ctx}}\), and \(\theta_{\text{cst}}\) in Eq.~\ref{equ:twop} and Eq.~\ref{equ:onep} are set to 1.0, 0.2, and 0.1, respectively. 
The number of object queries $N_q$ is set to 100.
The Transformer decoder is initialized with the parameters of an  object detector pre-trained on the MS-COCO dataset~\cite{lin2014microsoft}, excluding novel object categories. For object detection results, we retain object trajectories with an average classification score greater than 0.2 and filter bounding boxes using a threshold of 0.35.
The optimization process employs the AdamW algorithm~\cite{loshchilov2017decoupled} with an initial learning rate of 1e-4.
A multi-step decay schedule is applied at epochs 15, 20, and 25, reducing the learning rate by a factor of 0.1 at each step, and the model is trained for a total of 30 epochs. The batch size is set to 1, which means that only one video is processed at a time.
All experiments are conducted using a single NVIDIA GeForce RTX 4090 GPU.

\subsection{Comparison Results}
We compare our method with existing Open-VidVRD methods, including RePro~\cite{gao2023compositional}, UASAN~\cite{wu2024open}, OV-MMP~\cite{yang2024multi} and EOV-MMP~\cite{wang2024end}.
Notably, RePro, UASAN, and OV-MMP rely on trajectory detectors pre-trained on a close set. For a fair comparison, we exclude data from novel categories and retrain the trajectory detector to reproduce these methods.
Due to the lack of publicly available models or codes of RePro and UASAN on the SGDet task on the VidOR dataset, we does not show their results on the VidOR dataset.

Tab.~\ref{tab:sota-vidvrd} reports the evaluation results of our method and existing Open-VidVRD methods on the SGDet task on the VidVRD and VidOR datasets under both novel-split and all-split settings.
From Tab.~\ref{tab:sota-vidvrd}, we can draw the following observations:
(1) METOR achieves  improvements over contemporary models on almost all metrics on both datasets, especially  substancial gains in mAP and mAP$_o$.
This indicates that our method helps a lot improving the performance of object detection and relationship classification by  mutually enhancing the representations of objects and relationships.
(2)  Compared with the all-split, METOR achieves more significant improvements on the novel-split. For instance, in terms of mAP metric on  VidOR, our method surpasses the best competing approach by an absolute margin of 1.30\% (3.75\% vs. 2.45\%) and a relative margin of 53.06\% on the novel-split, while achieving an absolute margin of 1.24\% (12.32\% vs. 11.08\%) and a relative margin of 11.19\% on the all-split. 
This demonstrates that by leveraging the rich semantic knowledge in CLIP to capture object and relationship contexts to enhance feature representations, our method  improves the generalization ability to novel categories.

We also compare METOR with state-of-the-art visual language pre-trained models such as Video-LLMs, and the experimental results provided in {Supplementary Materials}.

\begin{table}
\centering
\begin{tabular}{cc|cc|cc}
\hline
\hline
\multirow{2}{*}{Enc} & \multirow{2}{*}{Itr}& \multicolumn{2}{c|}{Novel} & \multicolumn{2}{c}{All}\\
\cline{3-6}
 & &mAP&mAP$_o$&mAP&mAP$_o$\\
\hline
 & &11.64&29.38&24.33&48.27\\
\checkmark& &15.16&35.11&25.67&50.30\\
 &\checkmark&13.49&29.38&25.97&52.17\\
\checkmark &\checkmark&\textbf{16.43}&\textbf{38.56}&\textbf{27.45}&\textbf{55.09}\\
\hline
\hline
\end{tabular}
\caption{Performance of ablation study for the two modules in METOR on the VidVRD dataset. ``Enc" and ``Itr" denote the contextual refinement encoding module and the iterative enhancement module, respectively.}
\label{tab:abl_metor}
\end{table}

\subsection{Ablation Studies}
We conduct comprehensive ablation studies on the VidVRD dataset to assess the contribution of each component.

\noindent \textbf{Effectiveness of Different Modules.}
To evaluate the effectiveness of the contextual refinement encoding module (denoted as ``Enc"), we remove it and use the original CLIP encoder as its replacement for comparison.  To evaluate  the iterative enhancement module (denoted as ``Itr"), we remove it and design a non-mutual enhancement module as replacement.
Tab.~\ref{tab:abl_metor} shows the evaluation results where the consistent improvements in  all metrics highlight the effectiveness of the proposed modules in our method.
Specifically, incorporating the contextual refinement module boosts performance, particularly for novel categories, suggesting that contextual refinement improves generalization to real-world scenarios. 
Additionally, adding the iterative enhancement module improves both object and relationship detection, validating the benefits of mutual enhancement.

\noindent \textbf{Effectiveness of Contextual Embeddings.}
To evaluate the effectiveness of the contextual embeddings in the contextual refinement encoding module, we design several variants of METOR for comparison: (1) ``w/o CRE", removing contextual refinement encoding; (2) ``w/o CRQ", removing contextual refinement of object queries;  (3) ``w/o CRT", removing contextual refinement of text features.
As shown in Tab.~\ref{tab:ctx_tokens}, the results demonstrate that METOR outperforms all variants. Removing contextual refinement (whether applied to object queries, text features, or both) leads to significant  drops in performance, especially in the novel categories. These comparisons underscore the critical contribution of contextual refinement encoding to relationship detection.

\begin{table}
\centering
\begin{tabular}{c|cc|cc}
\hline
\hline
\multirow{2}{*}{Variant}& \multicolumn{2}{c|}{Novel} & \multicolumn{2}{c}{All}\\
\cline{2-5}
 &mAP&mAP$_o$&mAP&mAP$_o$\\
\hline
w/o CRE&13.49&29.38&25.97&52.17\\
w/o CRQ&15.32&35.19&26.25&53.30\\
w/o CRT&14.91&33.26&26.73&54.08\\
METOR&\textbf{16.43}&\textbf{38.56}&\textbf{27.45}&\textbf{55.09}\\
\hline
\hline
\end{tabular}
\caption{Performance of ablation study for the contextual refinement encoding on the VidVRD dataset.}
\label{tab:ctx_tokens}
\end{table}

\begin{table}
\centering
\begin{tabular}{c|cc|cc}
\hline
\hline
{Iteration}& \multicolumn{2}{c|}{Novel} & \multicolumn{2}{c}{All}\\
\cline{2-5}
 Number&mAP&mAP$_o$&mAP&mAP$_o$\\
\hline
0&15.16&35.11&25.67&50.30\\
1&16.04&37.82&26.88&53.76\\
2&\textbf{16.43}&\textbf{38.56}&\textbf{27.45}&{55.09}\\
3&16.25&38.28&27.39&\textbf{55.16}\\
\hline
\hline
\end{tabular}
\caption{Performance of model with different iteration numbers on the VidVRD dataset.}
\label{tab:iter_layers}
\end{table}

\noindent \textbf{Evaluation of the Number of Iteration.}
To evaluate the impact of the number of iterations in the iterative enhancement module, we start with a baseline model without the mutual enhancement, that is,  the iteration number is set to zero. We then gradually increase the number of iterations by adding more iterative enhancement layers and analyze their impact on the performance. As shown in Tab.~\ref{tab:iter_layers}, the results demonstrate that as the number of iterations increases, the  performance improves significantly, peak at two layers, and then decreases slightly.

\begin{figure}[t]
    \centering
    \includegraphics[width=0.98\linewidth]{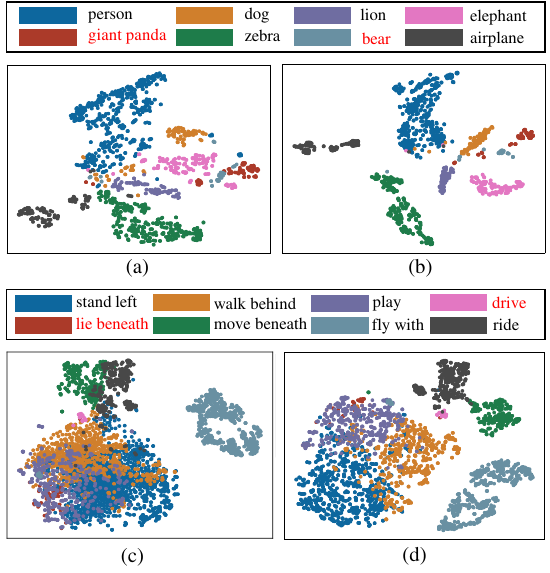}
    \caption{Qualitative results of feature distributions via T-SNE. (a) and (b) display the feature distribution of objects before and after mutual enhancement, while (c) and (d) show the feature distribution of relationships. The categories labeled in red font represent the novel categories.}
    \label{fig:feature}
    \end{figure}

\subsection{Qualitative Analysis}
To further evaluate the impact of the iterative mutual enhancement module on feature representation, we visualize the feature distributions of object and relationship categories before and after mutual enhancement. Specifically, we project the features onto a 2D plane using T-SNE~\cite{hinton2002stochastic}.  
As illustrated in Fig.~\ref{fig:feature}, the features after mutual enhancement exhibit better clustering. Features within the same category become more compact, while those between different categories are more clearly separated. This indicates that the mutual refinement process effectively enhances the feature discrimination of objects and relationships.  
More qualitative analysis are provided in the {Supplementary Materials}.

\section{Conclusion}
 In this paper, we propose METOR, a query-based unified framework that jointly models object detection and relationship classification. It is simple yet effective and can mutually enhance object detection and relationship classification by effectively exploiting their interdependence.
 Under this framework, we design an iterative enhancement module that alternately enhances the representations of objects and relationships by using each other's representation. 
 Additionally, we design a contextual refinement encoding module that extracts contexts for objects and relationships to refine the encoding of text features and object queries. 
 Extensive experimental results on the VidVRD and VidOR datasets demonstrate that our method achieves state-of-the-art performance.
 In the future, we will explore leveraging additional modalities such as audio and 3D information to support Open-VidVRD in a wider range of applications to enhance scene understanding in dynamic environments.

\section*{Acknowledgments}

This work was partially supported by the Shenzhen Science and Technology Program (Grant No. JCYJ20241202130548062).

\bibliographystyle{named}
\bibliography{ijcai25}

\end{document}